\documentclass[letterpaper]{article} % DO NOT CHANGE THIS
\usepackage[preprint]{aaai2027} % DO NOT CHANGE THIS
\usepackage[hyphens]{url} % DO NOT CHANGE THIS
\usepackage{graphicx} % DO NOT CHANGE THIS
\usepackage{natbib} % DO NOT CHANGE THIS AND DO NOT ADD OPTIONS
\usepackage{caption} % DO NOT CHANGE THIS AND DO NOT ADD OPTIONS
\usepackage{amsmath,amssymb}
\usepackage{booktabs}
\usepackage{tabularx}
\usepackage{array}
\usepackage{multirow}
\usepackage{xcolor}
\usepackage{tikz}

\usetikzlibrary{arrows.meta,positioning,calc,fit,backgrounds,shapes.geometric}

\newcommand{\Albilich}{\textsc{Albilich}}

\newcommand{\code}[1]{\texttt{#1}}

\colorlet{albProofDraw}{blue!45!black}
\colorlet{albProofFill}{blue!6!white}
\colorlet{albEvidenceDraw}{green!35!black}
\colorlet{albEvidenceFill}{green!7!white}
\colorlet{albControlDraw}{orange!55!black}
\colorlet{albControlFill}{orange!8!white}
\colorlet{albAdversaryDraw}{red!55!black}
\colorlet{albAdversaryFill}{red!7!white}
\colorlet{albUIDraw}{purple!55!black}
\colorlet{albUIFill}{purple!6!white}
\colorlet{albNeutralDraw}{black!65}
\colorlet{albNeutralFill}{black!3}
\tikzset{
  alb/base/.style={draw,rounded corners=2pt,align=center,font=\scriptsize,inner xsep=3pt,inner ysep=2.5pt},
  alb/proof/.style={alb/base,draw=albProofDraw,fill=albProofFill,minimum width=19mm,minimum height=6mm},
  alb/evidence/.style={alb/base,draw=albEvidenceDraw,fill=albEvidenceFill,minimum width=22mm,minimum height=6mm},
  alb/control/.style={alb/base,draw=albControlDraw,fill=albControlFill,minimum width=23mm,minimum height=6mm},
  alb/adversary/.style={alb/base,draw=albAdversaryDraw,fill=albAdversaryFill,minimum width=22mm,minimum height=6mm},
  alb/ui/.style={alb/base,draw=albUIDraw,fill=albUIFill,minimum width=24mm,minimum height=6mm},
  alb/flow/.style={-{Latex[length=1.8mm]},thick,draw=albNeutralDraw},
  alb/dashflow/.style={-{Latex[length=1.8mm]},thick,dashed,draw=albNeutralDraw},
  alb/sourceflow/.style={-{Latex[length=1.8mm]},thick,dashed,draw=albEvidenceDraw},
  alb/debtflow/.style={-{Latex[length=1.6mm]},densely dotted,thick,draw=albAdversaryDraw},
  alb/auditflow/.style={-{Latex[length=1.6mm]},densely dotted,thick,draw=albControlDraw},
  alb/observe/.style={-{Latex[length=1.6mm]},densely dotted,thick,draw=albUIDraw},
  alb/edgelabel/.style={font=\scriptsize,inner sep=1pt,fill=white,text=black}, 
  alb/strategyflow/.style={
  -{Latex[length=1.8mm]},
  thick,
  dashed,
  draw=albControlDraw},
}

\title{\Albilich: Steerable Proof-State Orchestration for LLM-Based Mathematical Research with CAS Integration}
\author{
Ting Gong\textsuperscript{\rm 1},
Michael Ruofan Zeng\textsuperscript{\rm 1},
Yong Yang\textsuperscript{\rm 2}
}

\affiliations{
\textsuperscript{\rm 1}University of Washington\\
Seattle, Washington, USA\\
\textsuperscript{\rm 2}Texas State University\\
San Marcos, Texas, USA\\
tgong2@uw.edu, zengrf@uw.edu, yang@txstate.edu
}

\begin{document}
\maketitle

\begin{abstract}
Large language models can contribute useful ideas to mathematical research, yet long-horizon proof attempts remain difficult to coordinate, evaluate, and reproduce. We present \Albilich{}, an open-source agentic harness for autoresearch in mathematics that combines long-horizon reasoning, computer algebra systems (CAS), literature retrieval, and persistent SQLite-based context management. 

We evaluate \Albilich{} on the RealMath benchmark \citep{ZhangEtAl2025RealMath} and on open problems in group theory from the Kourovka Notebook \cite{khukhro2026kourovka}. It solved 10/10 problems on RealMath with CAS and 9/10 with no CAS. On the Kourovka problems, \Albilich{} produced a counterexample to Problem 21.142 and a proof of a strengthening of Problem 20.2. An ablation on Problem 17.91 demonstrates $32.0\%$ token reduction when CAS is enabled. An ablation on Problem 21.142 demonstrates higher verifier-rejection rate and failure to synthesize proof routes in the absence of the advisor agent. These results support \Albilich{} as a human-steerable, CAS-boosted environment for scalable AI-assisted mathematical research.
\end{abstract}

\section{Introduction}
In recent years, many LLM-based proof systems and agentic harnesses have been developed to solve mathematical problems and contribute useful ideas to research-level questions. However, when a proof run continues for a long time, the research process can become opaque, even though failed approaches may contain mathematically useful information. For large proof constructions, a linear research system may repeatedly return to the same difficult ideas, leading to substantial costs in both search and logical deduction. Verification presents a further challenge: hypotheses may fail to match, dependencies may remain unchecked, and the system may produce arguments that appear plausible and convincing despite being false.

\Albilich{} is designed to address these problems and support a broader mathematical workflow. To our knowledge, it is among the first proof-harness agents organized around a persistent SQLite proof state, allowing it to retain proof graphs, failed routes, unresolved obligations, technical lemmas, and proof artifacts throughout a long run. Through a carefully designed user interface and MCP server, it supports continuous human observation and steering. A mathematician can inspect the current proof state, understand which routes have failed and which obligations remain, and redirect the research strategy when needed. This makes the process more transparent and places human--AI collaboration at the center of long-horizon mathematical research.

To address repeated cycling in a linear research process, \Albilich{} decomposes the main problem into more specific approaches and preserves each of them as an associated node in the proof graph. When the main researcher encounters a persistent obstruction, the PhD-advisor role can reformulate the current objective, reduce it to more tractable subproblems, or redirect the research strategy. The system also supports parallel computation across genuinely independent branches, including distinct proof methods and separate cases in classification problems. This architecture follows the philosophy of test-time scaling by allocating additional computation across structured and strategically selected research directions.

We introduce two kinds of tool calls via dedicated MCP servers. The first kind is computer algebra systems (CAS), including SageMath, Macaulay2, GAP, Singular, and Julia.  We demonstrate through an ablation study that the CAS tool call substantially reduces the cost of long-horizon mathematical research.  Specifically, the ablation on Problem 17.91 from the Kourovka Notebook shows a 32.0\% reduction in recorded token usage when CAS was enabled. Token consumption is recorded throughout each run and displayed in the user interface, allowing the mathematician to monitor computational cost together with mathematical progress. 

The second kind of tool call is literature retrieval. The \textit{literature researcher} role as MCP tool calls to \texttt{TheoremSearch} \cite{alexander2026theoremsearch} and \texttt{Matlas} \cite{ju2026matlas}. It records the exact source location, adapts notation, checks hypotheses, and states the precise form of the cited result required by the current proof strategy. In parallel, the adversarial researcher searches for obstructions and counterexamples while the researcher develops positive arguments. \textit{We will release an online MCP server for both kinds of tool calls in accordance with the release of this paper. }

We also make verification an independent part of the workflow. The researcher submits bounded proof \textit{dossiers} containing the target claim, its premises, the proposed inferences, and the supporting source or CAS artifacts. The PhD advisor may evaluate the current route and provide advice for overall strategy and high-leverage moves, but it does not provide a correctness verdict. A strict verifier then checks each dossier, including its deductions, hypotheses, citations, and computational evidence. When an argument is incomplete or an assumption fails to match, the verifier returns a precise proof \textit{debt}. After verification of each local statement, a separate integration verifier checks that the verified claims assemble into a correct proof of the original statement. This separation of research, strategy, local verification, integration, and refutation makes the acceptance process more explicit and auditable.

In summary, we make the following three main contributions. First, we introduce persistent SQLite proof state management that supports longer, more transparent, and auditable research runs. Second, we add a PhD-advisor role and graph-based problem decomposition to support test-time scaling, strategic redirection, and bootstrapped reductions when the main research process stalls. Third, we introduce MCP tool calls including CAS and theorem search tools and  demonstrate that the availability of CAS leads to a reduction in token usage. We have used \Albilich{} to resolve two problems in group theory, verified by human experts.

\begin{figure*}[htbp]
\centering
\begin{tikzpicture}[x=1cm,y=1cm]
  \node[alb/proof,text width=31mm] (research) at (-5.2,4.6)
    {Researcher\\proof, repair, source adaptation, CAS};
  \node[alb/evidence,text width=31mm] (literature) at (0,4.6)
    {Literature Reviewer\\primary-source cards and theorem interfaces};
  \node[alb/adversary,text width=31mm] (villain) at (5.2,4.6)
    {Adversarial Researcher\\stress-test, counterexample, obstruction};

  \node[alb/control,text width=39mm,minimum height=9mm] (gate) at (0,3.25)
    {Validated Patch Gate\\role, revision, schema, evidence};
  \node[alb/control,text width=69mm,minimum height=12mm] (state) at (0,1.9)
    {Canonical SQLite Proof State\\claims, routes, inferences, debts, sources, artifacts, budgets};

  \node[alb/control,text width=32mm] (scheduler) at (-4.7,0.45)
    {Deterministic Scheduler\\decisive obligation + next action};
  \node[alb/control,text width=32mm] (advisor) at (0,0.45)
    {PhD Advisor\\tactical steering + global synthesis};
  \node[alb/ui,text width=32mm] (dashboard) at (4.7,0.45)
    {Live Dashboard\\read-only graph, debts, runs, tokens};

  \node[alb/proof,text width=32mm] (strict) at (-4.7,-1.25)
    {Strict Verifier\\local proof, citation, and finite interfaces};
  \node[alb/proof,text width=32mm] (integration) at (0,-1.25)
    {Integration Verifier\\route sufficiency + root alignment};
  \node[alb/proof,text width=32mm] (status) at (4.7,-1.25)
    {Public Result State\\solved / partial / unresolved};
  \node[alb/ui,text width=32mm] (output) at (4.7,-2.55)
    {Status-Inert Outputs\\final proof, audit, revision, referee report};

  \draw[alb/flow] (research.south) -- (gate.north west);
  \draw[alb/sourceflow] (literature.south) -- (gate.north);
  \draw[alb/debtflow] (villain.south) -- (gate.north east);
  \draw[alb/flow] (gate) -- (state);

  \draw[alb/flow] (state.south west) -- (scheduler.north);
  \draw[alb/strategyflow] (state.south) -- (advisor.north);
  \draw[alb/observe] (state.south east) -- (dashboard.north);
  \draw[alb/strategyflow] (advisor.west) --
    node[alb/edgelabel,above] {advice} (scheduler.east);

  \draw[alb/flow] (scheduler.south) -- (strict.north);
  \draw[alb/flow] (strict) -- (integration);
  \draw[alb/flow] (integration) -- (status);
  \coordinate (debtA) at ($(strict.west)+(-0.95,0)$);
  \draw[alb/debtflow] (strict.west) -- (debtA)
      |- node[alb/edgelabel,pos=.28,left] {precise debt} (state.west);
  \draw[alb/observe] (status.north) -- ++(0,0.35) -| (dashboard.south);
  \draw[alb/observe] (status) -- (output);
\end{tikzpicture}
\caption{The \Albilich{} workflow.}
\label{fig:workflow}
\end{figure*}

\paragraph{Terminology.} We use the following terms throughout. The \emph{root} is the problem statement, fixed when a run begins. A \emph{claim} is a mathematical statement held in the proof state, and an \emph{inference} is one proposed deduction from premise claims to a conclusion. A \emph{route} is a chain of inferences advanced as a proof of a claim, and is \emph{sufficient} when its final inference entails its conclusion from the stated premises. A \emph{debt} records an unmet obligation, such as a gap or a missing hypothesis, against the object that owes it; the debts together form the \emph{debt ledger}. A \emph{patch} is an attributable edit proposed by one agent session. The \emph{proof spine} of the root is the set of claims reachable from it through integrated routes.

\paragraph{Acknoledgements.} This project is a part of the UW Math AI Lab. Ting Gong thanks Yuan Lu for advice and for testing \Albilich{}, Lu Qi for helpful questions and interest in the project, Shitan Xu and Yu Shen for testing the program, and his advisor, Max Lieblich, for his interest and support. He further thanks Bin Dong for helpful conversations. Michael R. Zeng thanks Jarod Alper, Vasily Ilin, and Gergely Bérczi for many helpful discussions. 

\section{Related Work}

Several recent systems use language models to support research-level mathematics. Rethlas combines informal proof search with theorem retrieval, while Archon converts selected informal arguments into machine-checkable Lean~4 code through decomposition, refinement, and automated proof synthesis \cite{ju2026rethlas}. Danus coordinates parallel workers through a shared fact graph. A stateless verifier checks proposed claims before adding them to the graph \cite{liu2026danus}. QED separates proof planning, proof construction, and verification. Its design aims to reduce context contamination, citation errors, unstable plans, diffuse verification, and changes to the target problem \cite{an2026qed}. Aletheia uses repeated cycles of generation, verification, revision, and tool use for long-horizon mathematical research \cite{feng2026aletheia}. AI Co-Mathematician instead provides an interactive workspace for literature search, computation, theorem proving, and theory development \cite{zheng2026comathematician}.

Other systems focus on formal verification or executable evaluation. Lean~4 provides kernel-checked verification through a small trusted core \cite{moura2021lean4}. LeanDojo studies retrieval-augmented theorem proving in Lean \cite{yang2023leandojo}, while DeepSeek-Prover uses proof-assistant feedback and search to improve formal proof generation \cite{xin2024deepseekprover}. FunSearch takes a different approach by evaluating model-generated programs through execution \cite{romeraparedes2024funsearch}. 

\Albilich{} primarily addresses the informal stage of mathematical research. It records candidate statements, proof routes, source interfaces, obstructions, counterexamples, and partial results before formalization. \textit{The status \code{formally\_verified} is reserved for claims supported by evidence from a formal backend, the implementation of which we defer to the next version.} \Albilich{} also integrates with TheoremSearch for literature retrieval \cite{alexander2026theoremsearch}. Retrieved candidates remain untrusted until a literature researcher checks the primary source, matches its hypotheses and definitions, and verifies the implication needed in the current proof.

Danus is the closest system in its use of a verified fact graph and global coordination \cite{liu2026danus}. \Albilich{} differs in four respects. It retains proposed, rejected, and superseded objects in a typed event-sourced state. It represents unresolved obligations as explicit proof debts. It separates local claim verification from root-level proof integration. It also reopens an integrated route when changes to its dependencies invalidate the accepted proof spine. QED likewise separates planning, proving, and verification \cite{an2026qed}. In \Albilich{}, however, acceptance authority belongs to designated roles rather than to fixed positions in a pipeline. ProofCouncil pairs an author with a critic that is reset between rounds and supplements this loop with a CAS node \cite{schmitt2026proofcouncil}. Its critic evaluates a manuscript as a whole. By contrast, the \Albilich{} integration verifier checks whether locally accepted claims and inferences form a sufficient route to the immutable root statement.

\begin{table}[htbp]
\centering\small
\caption{Architectural comparison. }
\label{tab:related}
\setlength{\tabcolsep}{2.5pt}
\begin{tabular}{@{}lcccc@{}}
\toprule
Capability & Danus & QED & ProofC. & \Albilich{} \\
\midrule
Retains rejected/subsumed claims& -- & -- & -- & \checkmark \\
Explicit debt/obligation ledger & -- & -- & -- & \checkmark \\
Local \emph{vs} root integration check & -- & partial & -- & \checkmark \\
Role-authorized state transitions & -- & partial & -- & \checkmark \\
Dependency-triggered invalidation & -- & -- & -- & \checkmark \\
Source/CAS evidence lineage & partial & \checkmark & \checkmark & \checkmark \\
Human steering & \checkmark & -- & -- & \checkmark \\
\bottomrule
\end{tabular}
\end{table}
\section{Proof-State Architecture}

\Albilich{} represents a proof attempt as a versioned graph with attached evidence, open obligations, and a record of state transitions. Agent sessions receive role-specific views of this state. They may modify it only by submitting validated patches.

\subsection{Proof-State Representation}

Let $q_\star$ denote the immutable root statement. At revision $t$, the proof state consists of
\begin{equation}
\begin{aligned}
S_t
  &=
  (q_\star,G_t,\mathcal D_t,\mathcal L_t,\mathcal A_t,
   \mathcal P_t,\mathcal H_t,\mathcal M_t),\\
G_t
  &=
  (C_t\mathbin{\dot\cup}R_t\mathbin{\dot\cup}I_t,\Gamma_t).
\end{aligned}
\label{eq:state}
\end{equation}
Here, $C_t$, $R_t$, and $I_t$ are the sets of claims, proof routes, and inferences. The typed edges $\Gamma_t$ record logical dependencies among them. A route consists of claims and inferences intended to establish the root statement or a stated reduction of it.

The remaining components store information associated with the proof graph. The debt ledger $\mathcal D_t$ records unresolved gaps and missing interfaces. The library $\mathcal L_t$ contains retrieval records with source locations, hypotheses, and theorem statements. The artifact store $\mathcal A_t$ contains proof arguments, source adaptations, CAS outputs, and verification reports. The sets $\mathcal P_t$ and $\mathcal H_t$ record accepted patches and state-transition events. Resource measurements are stored in $\mathcal M_t$.

Validation and lifecycle status are recorded separately. A claim may be locally verified without belonging to an integrated proof route. Conversely, an active claim may remain unverified. Superseded claims remain in the history but are removed from the active search frontier.

\subsection{Controlled State Updates}

A session proposes a patch
\begin{equation}
\pi=(\rho,b,X,\Omega,\eta),
\label{eq:patch}
\end{equation}
where $\rho$ identifies the submitting role, $b$ is the revision on which the session operated, $X$ lists the affected objects, $\Omega$ contains the requested operations, and $\eta$ contains the supporting evidence. The state transition is
\begin{equation}
S_{t+1}
=
\begin{cases}
\operatorname{Apply}(S_t,\pi),
  & \operatorname{Valid}(S_t,\pi),\\
\operatorname{LogReject}(S_t,\pi),
  & \text{otherwise}.
\end{cases}
\label{eq:transition}
\end{equation}

Patch validation checks that the submitting role is authorized to request each operation. It also checks that the base revision is compatible with the current state, all referenced objects exist, the patch satisfies its schema, and the required evidence is present. Application is atomic. An invalid patch does not modify the mathematical state, although its rejection and diagnostic are recorded in the event history.

A patch produced from an earlier revision may be reconsidered only when its target and relevant dependency closure have not changed. Otherwise, the result must be regenerated or reverified against the current state. 

\begin{table*}[htbp]
\centering
\caption{Roles in \Albilich{}. Mathematical status may be changed only by the three checking roles.}
\label{tab:roles}
\scriptsize
\setlength{\tabcolsep}{3pt}
\renewcommand{\arraystretch}{1.03}
\begin{tabular}{@{}p{0.245\textwidth}p{0.595\textwidth}p{0.105\textwidth}@{}}
\hline
Role & Function & Authority \\
\hline
Researcher
& Constructs proofs, repairs routes, adapts sources, runs CAS, and prepares proof dossiers.
& None \\

Adversarial researcher
& Stress-tests claims, identifies obstructions, and proposes counterexamples.
& None \\

Literature researcher
& Records primary-source statements, hypotheses, and source adaptations.
& None \\

PhD advisor
& Synthesizes evidence, identifies bottlenecks, and redirects routes and work modes.
& None \\

Strict verifier
& Checks bounded proof, source, and CAS packets.
& Verify \\

Integration verifier
& Checks route sufficiency and alignment with the immutable root.
& Integrate \\

Counterexample validator
& Checks that a proposed counterexample satisfies the hypotheses and refutes the claim.
& Refute \\

Writer
& Renders accepted proofs or explicitly partial results.
& None \\

Scheduler
& Selects the next action from the proof state.
& None \\
\hline
\end{tabular}
\end{table*}

\subsection{Agent Interaction via SQLite}

\Albilich{} stores the canonical proof state in a versioned SQLite database. At each iteration, the scheduler selects a role and a target object from the current state. The target may be a claim, inference, route, proof debt, source interface, or verification request.

The selected session receives a snapshot of the state at revision $b$. Its input contains the assigned target, the relevant dependency subgraph, active debts on that subgraph, and the evidence needed for the task. Historical material outside this subgraph remains in the database but is omitted from the session context. This limits prompt size without discarding earlier work.

A session cannot edit the database directly. It returns a structured patch against revision $b$. Patch validation and application follow Equation~\eqref{eq:transition}. For a fixed state and configuration, the scheduler's choice is deterministic. The model output produced by the selected session remains stochastic.

\subsection{Verification and Integration}
We state the conservative acceptance criteria for a research run and describe how local verification and root-level integration are separated to prevent premature acceptance.

\paragraph{Authority of roles.}
Research roles may introduce or revise claims, routes, arguments, source records, and candidate counterexamples. They cannot verify, refute, or integrate them. The writer and dashboard also have no authority over mathematical status.

The strict verifier may assign \code{informally\_verified} after checking a bounded argument and its premises. The status \code{formally\_verified} requires evidence accepted by a formal backend. Refutation requires acceptance through the designated verification or counterexample-validation procedure. Only the integration verifier may mark a route as integrated.

This separation prevents an agent that constructs an argument from also certifying that argument. A repair may add evidence or replace an inference, but any associated debt remains open until the appropriate verifier accepts the change.

\paragraph{Local verification.}
Local verification concerns a single claim or inference. The verifier checks the stated premises, quantifiers, hypothesis propagation, case coverage, and any source or computational interfaces used by the argument. Local acceptance establishes only the submitted claim. It does not establish that the claim contributes to a proof of $q_\star$.

\paragraph{Route integration.}
The integration verifier checks whether locally accepted claims and inferences form a complete route to the root statement. A route may be integrated only if its terminal inference is verified, every premise in its dependency closure is verified, and no blocking debt remains on that closure. The integration report must also identify the relation between the route's conclusion and $q_\star$ as exact, equivalent, or stronger. For a stronger conclusion, the implication to $q_\star$ must be checked explicitly. The completion criterion is straightforward: all local verifications must pass, and an integrated proof route must be present.

\paragraph{Reconciliation.}
The system rechecks an integrated route whenever one of its dependencies changes. Integration is withdrawn if a required claim loses its verified status, an accepted inference is replaced, a blocking debt is introduced, or the route no longer establishes the recorded relation to the root. The affected route then returns to verification or repair. This process is independent from generated artifacts. 

\subsection{Task Scheduling}

The scheduler constructs tasks from unresolved items in the active proof graph. These items include unproved premises, unchecked inferences, open proof debts, claims awaiting verification, and routes awaiting integration. We refer to them collectively as \emph{obligations}.

The current implementation uses a rule-based priority function. It gives preference to obligations that block an active route to the root statement. The ranking also accounts for whether an obligation is shared by several routes, whether another session is already working on it, and the estimated cost of the required research or verification step. The run configuration fixes the priority rules and tie-breaking order.

The scheduler may assign independent obligations to parallel sessions. Whenever possible, concurrent sessions receive distinct targets so that they do not reproduce the same work. Research and verification draw from separate budget allocations. This prevents speculative search from exhausting the resources reserved for checking completed arguments.

The PhD advisor is a long-horizon planning role. It is invoked after a configured number of steps without an accepted root-relevant change, after repeated rejection of the same proposed inference, when several routes encounter the same debt, or when a central claim is refuted. The advisor may propose a new reduction, reprioritize active routes, pause an unproductive branch, or recommend a budget for a branch. These recommendations are recorded as planning decisions. The advisor cannot verify claims, close debts, or integrate routes.

Each subsequent session receives only the advisor decisions relevant to its assigned target. The full planning history remains available in the proof state for later inspection.

\subsection{MCP Tool Calls: Literature Search and Computer Algebra}

We now discuss the implementation details of the MCP tools calls available to the agents under various roles. 

\paragraph{Literature sources.}
External theorem-search services, including TheoremSearch, return candidate references \cite{alexander2026theoremsearch}. Search results are treated as retrieval aids rather than mathematical evidence. Before a source can support a claim, the literature researcher records the exact theorem statement, its location in the primary source, its hypotheses, the translation to the notation of the current problem, and the implication required by the proof.

The strict verifier checks this source record together with the proposed local deduction. A retrieved theorem cannot verify a claim unless its hypotheses and its relation to the target claim have both been established.

\paragraph{Computer algebra systems (CAS).}

Computer-algebra access is mediated by a Model Context Protocol
server~\cite{mcp2025} as tool calls for the researcher and adversarial
roles. Sessions are asynchronous, so a role starts a computation,
continues reasoning, and polls for the result. The server routes each session to
one of four backends, SageMath~\cite{sagemath}, GAP~\cite{GAP4},
Macaulay2~\cite{M2}, and Singular~\cite{DGPS}. Each session runs under a
concurrency cap, a per-session timeout, and an output-size limit, and its
artifact records the mathematical question, the finite scope, the exact code, the
output, and the backend version. The strict verifier reads that artifact and does
not re-execute the computation, so a CAS-supported claim carries the trust level
of its transcript.

A CAS session is assigned a mathematical question and an explicit computational scope. Its artifact records the software and version, input data, code or commands, output, and the conclusion drawn from the computation. The code and output are retained with the run.

A finite computation may establish an explicitly finite claim or provide a counterexample. It does not by itself establish a universal statement. Any such use requires a separate argument showing that the finite computation covers all relevant cases or that the computed statement implies the desired conclusion.

The strict verifier checks the mathematical interface between the CAS artifact and the claim that uses it. In the current implementation, the verifier does not independently rerun the computation. The archived code, inputs, and outputs therefore support reproducibility, while acceptance of the mathematical conclusion still depends on verification of the reduction to the reported computation.

\section{Experiments}
\label{sec:experiments}

We evaluate \Albilich{} along four directions. First, we test whether the complete workflow can produce verified results on research-level mathematical questions. Second, we study whether CAS work modes can reduce the token cost of a long proof attempt. Third, we report two longer finite-group-theory case studies in which the system was asked to address problems from the Kourovka Notebook \cite{khukhro2026kourovka}. Finally, we conduct an ablation of the PhD-advisor role as a mechanism for long-horizon reasoning. In the CAS ablation, the two conditions use the same problem statement, model, reasoning effort, one-hour attempt limit, and number of recorded child runs, while differing only in whether CAS work mode is available. In the PhD-advisor ablation, the two conditions begin from fresh proof states at the same frozen source revision and use the same problem statement, model, reasoning effort, research mode, search policy, parallelism, and token budget, while differing only in whether the PhD advisor may synthesize the proof state and redirect subsequent research. We separate the system's internal verdict from external mathematical evaluation throughout. All reported runs use \texttt{gpt-5.6-sol} with \texttt{xhigh} reasoning.

\begin{table}[htbp]
\centering
\caption{Summary of the reported experiments. Token counts include cached input. ``Final'' denotes Albilich's internal
\texttt{solved\_final} state.}
\label{tab:experiment-summary}
\small
\begin{tabular}{lrrl}
\hline
Experiment & Runs & Tokens & Outcome \\
\hline
RealMath Math\_arXiv & 10 problems & 28.10M & 10 final \\
17.91, CAS on & 20 & 6.78M & certified partial \\
17.91, CAS off & 20 & 9.98M & certified partial \\
21.142, advisor on & 80 & 29.68M & exact solution \\
21.142, advisor off & 110 & 54.90M & unsolved at stop \\
20.2, advisor on & 23 & 6.71M & explicit witness \\
20.2, advisor off & 18 & 6.31M & same witness \\
\hline
\end{tabular}
\end{table}

\subsection{RealMath Benchmark}

We evaluated \Albilich{} on ten problems from the Math\_arXiv portion of RealMath \cite{ZhangEtAl2025RealMath}, with CAS enabled in every run. All ten attempts reached \texttt{solved\_final}. Human comparison with the supplied references found nine clear matches; for Problem 08, the equivalence between \Albilich{}'s convolution formula and the reference expression remained unresolved. The runs used approximately 28.1 million gross tokens in total. This small sample tests completion of the proof-state, CAS, verification, integration, and writing workflow, and does not estimate performance on the full benchmark.

\subsection{CAS Ablation}

We use Kourovka Problem 17.91 to study the effect of the CAS work mode. We conducted two one-hour runs on an identical problem statement, using the same model and reasoning effort, with twenty recorded child sessions in each condition. CAS mode was enabled in one run and globally disabled in the other. The active backend compute was similar: $5244.145$ seconds with CAS and $5410.717$ seconds without CAS.

\begin{table}[t]
\centering
\caption{CAS ablation on Problem 17.91. Token counts include cached input.}
\label{tab:cas-ablation}
\small
\begin{tabular}{lrrr}
\hline
Condition & Input & Output & Total \\
\hline
CAS on  & 6.590M & 0.194M & 6.784M \\
CAS off & 9.787M & 0.195M & 9.982M \\
\hline
\end{tabular}
\end{table}

The CAS-enabled run recorded $32.0\%$ fewer gross tokens. The reduction was concentrated in input tokens, while output-token use was essentially unchanged. Neither run solved the root problem. This paired experiment provides preliminary evidence that CAS-assisted work modes can reduce the token cost of a long mathematical research attempt. 

\subsection{Case Studies: Open Problems in Group Theory}

We conduct two case studies on our attempts using \Albilich{} to attack open problems in group theory, selected from the Kourovka notebook \citep{khukhro2026kourovka}.

\paragraph{Kourovka Problem 21.142.}
For fixed distinct primes $p$ and $q$, Problem 21.142 asks whether every finite group embeds into a finite group invariably generated by elements of orders $p$ and $q$. \Albilich{} reported and internally certified a negative answer: for some $m\geq 9$, depending on $p$ and $q$, the alternating group $A_m$ cannot embed into any such group.

The accepted route reduces a minimal hypothetical host to a group between $S^t$ and $\operatorname{Aut}(S)\wr\operatorname{Sym}(t)$, where $S$ contains $A_m$. A terminal simple-factor obstruction then places suitable independent conjugates of the two generators in a common proper subgroup, contradicting invariable generation. The run used 80 child sessions, 5 hours 51 minutes of lifecycle time, 6.81 hours of active backend compute, and 29,684,464 gross tokens. Ten of eleven claims were verified, and one route concluding the root was integrated. Residual ledger debts remain preserved in the public report.

\paragraph{Kourovka Problem 20.2.}
Problem 20.2 asks whether a nonabelian simple group of Lie type can be totally $3$-closed. The archived run certified the explicit answer $G=\operatorname{PSL}_2(7)$. The proof reduces arbitrary faithful actions to unions of at most two transitive coset actions, verifies the resulting finite configurations with GAP, and uses a synchronization argument to pass to arbitrary faithful actions. The advisor-on run used 23 child sessions, 1 hour 16 minutes of lifecycle time, and 6,714,757 tokens.

A separate advisor-off run also certified the same witness using 18 child sessions, 1 hour 47 minutes of lifecycle time, and 6,306,998 tokens. Its proof relied on an exhaustive computational certificate rather than the theory-based direct argument produced by the advisor-on run. The advisor-off run took approximately 31 minutes longer, while its gross token use was comparable. This comparison suggests that, on this instance, the advisor mainly affected the mathematical character and organization of the proof, steering the system toward a more conceptual argument.

Further Albilich-assisted runs moved from the existence question toward a an almost complete family-level classification for $\operatorname{PSL}_n(q)$. They classify the
$\operatorname{PSL}_2(q)$, $\operatorname{PSL}_3(q)$, and
$\operatorname{PSL}_4(q)$ families and establish broad
higher-rank obstructions. The only remaining family is
$\operatorname{PSL}_n(2)$ for $n\geq 5$. These follow-up results combine structural reductions with exact finite certificates and are reported as a broader research program.

\subsection{PhD-Advisor Ablation}
\label{sec:advisor-ablation}

We evaluate the PhD advisor as a mechanism for test-time scaling through a matched-model ablation on Problem 21.142. The advisor-off arm began from a fresh proof state with the identical problem file. Both conditions used \texttt{gpt-5.6-sol} with \texttt{xhigh} reasoning, an 80-million-token budget, a 12-million-token verification reserve, a maximum reduction depth of four, and the \texttt{full\_proof\_first} completion policy. The only intended intervention was \texttt{ALBILICH\_ADVISOR\_ENABLED=0}; when the advisor was disabled, its scheduler opportunities were reassigned to other eligible actions.

\begin{table}[t]
\centering
\caption{PhD-advisor ablation on Problem 21.142. Parentheses give the number of root-concluding routes. Raw tokens include cached input, while charged tokens exclude cached input. C/V denotes total/verified claims.}
\label{tab:advisor-ablation}
\scriptsize
\setlength{\tabcolsep}{2.3pt}
\resizebox{\columnwidth}{!}{
\begin{tabular}{lcccccc}
\hline
Advisor & Root & Runs & Compute & Tokens & C/V & Reject. \\
\hline
Enabled  & solved (1)   & 80  & 6.81 h  & 29.7/6.00M & 11/10 & 1\% \\
Disabled & unsolved (0) & 110 & 10.44 h & 54.9/9.67M & 20/18 & 8\% \\
\hline
\end{tabular}
}
\end{table}

The advisor-off branch remained unsolved when it was stopped after 10.44 hours of active compute, or $1.53$ times the advisor-on solve time. It also used $1.85$ times the raw tokens and $1.61$ times the charged tokens. Since the run was stopped before its full two-times allowance, the comparison establishes only that it had not solved the problem by the stopping point, despite receiving substantially more compute.

The advisor-off branch nevertheless produced 18 verified claims, compared with 10 in the baseline, and accumulated 16 integrated routes. None concluded the root. The advisor-on run instead assembled one connected proof spine through the projective linear, symplectic, unitary, orthogonal, alternating, and terminal simple-factor cases.

The patch-rejection rate was also higher without the advisor, increasing from $1\%$ to $8\%$. This is consistent with a larger proportion of unsuccessful or poorly aligned state updates in the advisor-off run, even though the advisor-off system produced substantially more verified claims.

On this problem, the PhD advisor appears to contribute mainly through global direction, route selection, and proof assembly.

\section{Limitations and Conclusion}
\label{sec:conclusion}

\paragraph{Broader informal trials.}
One of the authors also tested \Albilich{} on graduate-level group theory and representation-theory projects. It completed two of three master's- or early-PhD-level problems, repaired a gap in one associated paper, and did not solve the third problem, which had also remained unresolved by the students. On several substantially harder open problems, it obtained no complete solution: brute-force search performed poorly on a codegree-set construction problem, while an open character-vertex problem received only preliminary ideas. Even in these unsuccessful runs, \Albilich{} produced explicit partial results, intermediate claims, and remaining proof debts that could be inspected and separately verified. In paper-audit trials, it quickly identified the known flaw in a published argument and produced detailed reports on two undergraduate research drafts, although some comments were overly scrupulous. These trials are anecdotal and are not counted as benchmark results.

\paragraph{Limitations}
The quantitative evaluation is preliminary and should be expanded through broader and repeated experiments. Internal informal verification remains distinct from formal verification, independent peer review, and external benchmark grading.

\paragraph{Conclusions}
Within these limits, the experiments illustrate the intended role of \Albilich{}. The system preserves long mathematical states, combines literature and computation with proof construction, records explicit debts when a theorem remains open, and exposes the computational cost of each run. The CAS comparison suggests that computational artifacts can reduce repeated context use. The Kourovka case studies show how local results and finite certificates can be assembled into readable research outputs. The advisor ablation further suggests that additional inference alone does not guarantee root progress: on Problem 21.142, the advisor-off arm produced more verified local mathematics while failing to assemble a route concluding the theorem.

\paragraph{Use of generative AI.}
Generative AI was used as the experimental subject of this work and for language editing of author-written text. The authors reviewed and approved the complete manuscript and remain responsible for its mathematical claims, citations, figures, tables, and conclusions.

\bibliography{albilich_references}

@misc{an2026qed,
  author        = {Chenyang An and Qihao Ye and Minghao Pan and Jiayaun Zhang},
  title         = {{QED}: An Open-Source Multi-Agent System for Generating Mathematical Proofs on Open Problems},
  year          = {2026},
  eprint        = {2604.24021},
  archivePrefix = {arXiv},
  primaryClass  = {cs.AI}
}

@misc{ju2026rethlas,
  author        = {Haocheng Ju and Guoxiong Gao and Jiedong Jiang and Bin Wu and Zeming Sun and Shurui Liu and Leheng Chen and Yutong Wang and Yuefeng Wang and Zichen Wang and Wanyi He and Peihao Wu and Liang Xiao and Ruochuan Liu and Bryan Dai and Bin Dong},
  title         = {Automated Conjecture Resolution with Formal Verification},
  year          = {2026},
  eprint        = {2604.03789},
  archivePrefix = {arXiv},
  primaryClass  = {cs.LG}
}

@misc{yang2023leandojo,
  author        = {Kaiyu Yang and Aidan M. Swope and Alex Gu and Rahul Chalamala and Peiyang Song and Shixing Yu and Saad Godil and Ryan Prenger and Anima Anandkumar},
  title         = {{LeanDojo}: Theorem Proving with Retrieval-Augmented Language Models},
  year          = {2023},
  eprint        = {2306.15626},
  archivePrefix = {arXiv},
  primaryClass  = {cs.AI}
}

@misc{xin2024deepseekprover,
  author        = {Huajian Xin and Z. Z. Ren and Junxiao Song and Zhihong Shao and Wanjia Zhao and Haocheng Wang and Bo Liu and Liyue Zhang and Xuan Lu and Qiushi Du and Wenjun Gao and Qihao Zhu and Dejian Yang and Zhibin Gou and Z. F. Wu and Fuli Luo and Chong Ruan},
  title         = {{DeepSeek-Prover-V1.5}: Harnessing Proof Assistant Feedback for Reinforcement Learning and Monte-Carlo Tree Search},
  year          = {2024},
  eprint        = {2408.08152},
  archivePrefix = {arXiv},
  primaryClass  = {cs.CL}
}

@inproceedings{moura2021lean4,
  author    = {Leonardo de Moura and Sebastian Ullrich},
  title     = {The {Lean 4} Theorem Prover and Programming Language},
  booktitle = {Automated Deduction -- CADE 28},
  series    = {Lecture Notes in Computer Science},
  volume    = {12699},
  pages     = {625--635},
  publisher = {Springer},
  year      = {2021}
}

@misc{zheng2026comathematician,
  author        = {Daniel Zheng and Ingrid von Glehn and Yori Zwols and Iuliya Beloshapka and Lars Buesing and Daniel M. Roy and Martin Wattenberg and Bogdan Georgiev and Tatiana Schmidt and Andrew Cowie and Fernanda Viegas and Dimitri Kanevsky and Vineet Kahlon and Hartmut Maennel and Sophia Alj and George Holland and Alex Davies and Pushmeet Kohli},
  title         = {{AI Co-Mathematician}: Accelerating Mathematicians with Agentic {AI}},
  year          = {2026},
  eprint        = {2605.06651},
  archivePrefix = {arXiv},
  primaryClass  = {cs.AI}
}

@misc{feng2026aletheia,
  author        = {Tony Feng and Trieu H. Trinh and Garrett Bingham and Dawsen Hwang and Yuri Chervonyi and Junehyuk Jung and Joonkyung Lee and Carlo Pagano and Sang-hyun Kim and Federico Pasqualotto and Sergei Gukov and Jonathan N. Lee and Junsu Kim and Kaiying Hou and Golnaz Ghiasi and Yi Tay and Yaguang Li and Chenkai Kuang and Yuan Liu and Hanzhao Lin and Evan Zheran Liu and Nigamaa Nayakanti and Xiaomeng Yang and Heng-tze Cheng and Demis Hassabis and Koray Kavukcuoglu and Quoc V. Le and Thang Luong},
  title         = {Towards Autonomous Mathematics Research},
  year          = {2026},
  eprint        = {2602.10177},
  archivePrefix = {arXiv},
  primaryClass  = {cs.AI}
}

@article{romeraparedes2024funsearch,
  author  = {Bernardino Romera-Paredes and Mohammadamin Barekatain and Alexander Novikov and Matej Balog and M. Pawan Kumar and others},
  title   = {Mathematical Discoveries from Program Search with Large Language Models},
  journal = {Nature},
  volume  = {625},
  pages   = {468--475},
  year    = {2024},
  doi     = {10.1038/s41586-023-06924-6}
}

@article{liu2026danus,
  author        = {Jihao Liu and Guoxiong Gao and Zeming Sun and Bin Wu
                   and Shurui Liu and Jiedong Jiang and Haocheng Ju
                   and Leheng Chen and Ronnie Cheng and Xiping Zhang
                   and Bin Dong},
  title         = {Danus: Orchestrating Mathematical Reasoning Agents
                   with Fact-Graph Memory},
  journal       = {arXiv preprint arXiv:2607.06447},
  year          = {2026},
  eprint        = {2607.06447},
  archivePrefix = {arXiv},
  url           = {https://arxiv.org/abs/2607.06447}
}

@book{khukhro2026kourovka,
  editor    = {Khukhro, Evgeny I. and Mazurov, Victor D.},
  title     = {Unsolved Problems in Group Theory: The Kourovka Notebook},
  edition   = {21st},
  number    = {21},
  publisher = {Sobolev Institute of Mathematics},
  address   = {Novosibirsk},
  year      = {2026},
  note      = {July 2026 update}
}

@article{alexander2026theoremsearch,
  author        = {Luke Alexander and Eric Leonen and Sophie Szeto
                   and Artemii Remizov and Ignacio Tejeda
                   and Giovanni Inchiostro and Vasily Ilin},
  title         = {Semantic Search over 9 Million Mathematical Theorems},
  journal       = {arXiv preprint arXiv:2602.05216},
  year          = {2026},
  eprint        = {2602.05216},
  archivePrefix = {arXiv},
  doi           = {10.48550/arXiv.2602.05216},
  url           = {https://arxiv.org/abs/2602.05216}
}

@inproceedings{ZhangEtAl2025RealMath,
  author    = {Jie Zhang and Cezara Petrui and Kristina Nikoli{\'c}
               and Florian Tram{\`e}r},
  title     = {{RealMath}: A Continuous Benchmark for Evaluating
               Language Models on Research-Level Mathematics},
  booktitle = {Advances in Neural Information Processing Systems},
  volume    = {38},
  year      = {2025}
}

@misc{ju2026matlas,
  title         = {{Matlas}: A Semantic Search Engine for Mathematics},
  author        = {Haocheng Ju and Leheng Chen and Peihao Wu and Bryan Dai and Bin Dong},
  year          = {2026},
  eprint        = {2604.17484},
  archivePrefix = {arXiv},
  primaryClass  = {cs.IR}
}

@misc{schmitt2026proofcouncil,
  title         = {{ProofCouncil}: An {LLM} Agent for Solving Open Mathematical Problems},
  author        = {Johannes Schmitt and Tim Gehrunger and Jasper Dekoninck and Gergely B{\'e}rczi and Uri Kreitner and Liam Price and David Holmes},
  year          = {2026},
  eprint        = {2607.09474},
  archivePrefix = {arXiv},
  primaryClass  = {cs.AI}
}

@misc{mcp2025,
  title        = {Model Context Protocol Specification},
  author       = {{Anthropic}},
  year         = {2025},
  note         = {Revision 2025-11-25},
  howpublished = {\url{https://modelcontextprotocol.io/specification}}
}

@manual{sagemath,
  organization = {The Sage Developers},
  title        = {{S}age{M}ath, the {S}age {M}athematics {S}oftware {S}ystem},
  year         = {2026},
  note         = {\url{https://www.sagemath.org}}
}

@manual{GAP4,
  organization = {The GAP~Group},
  title        = {{GAP -- Groups, Algorithms, and Programming}},
  year         = {2026},
  note         = {\url{https://www.gap-system.org}}
}

@misc{M2,
  author       = {Grayson, Daniel R. and Stillman, Michael E.},
  title        = {Macaulay2, a software system for research in algebraic geometry},
  year = {2026},
  howpublished = {\url{http://www2.macaulay2.com}}
}

@misc{DGPS,
  author       = {Decker, Wolfram and Greuel, Gert-Martin and Pfister, Gerhard
                  and Sch\"onemann, Hans},
  title        = {{\sc Singular} --- {A} computer algebra system for polynomial
                  computations},
  year         = {2024},
  howpublished = {\url{http://www.singular.uni-kl.de}}
}

\end{document}